\documentclass{article}

\usepackage[preprint, nonatbib]{nips}

\usepackage[utf8]{inputenc} 
\usepackage[T1]{fontenc}    
\usepackage{hyperref}       
\usepackage{url}            
\usepackage{booktabs}       
\usepackage{amsfonts}       
\usepackage{nicefrac}       
\usepackage{microtype}      

\usepackage{authblk}
\usepackage{amsmath, amssymb}
\usepackage{graphicx, wrapfig}
\usepackage{multirow}
\usepackage{caption, subcaption}
\usepackage[round]{natbib}

\usepackage{amsmath,amsfonts,bm}

\def\eqref#1{equation~\ref{#1}}

\def\1{\bm{1}}

\def\rvc{{\mathbf{c}}}

\def\rvx{{\mathbf{x}}}

\def\rvz{{\mathbf{z}}}

\def\ervz{{\textnormal{z}}}

\def\vc{{\bm{c}}}

\def\vx{{\bm{x}}}

\def\vz{{\bm{z}}}

\def\evz{{z}}

\DeclareMathAlphabet{\mathsfit}{\encodingdefault}{\sfdefault}{m}{sl}
\SetMathAlphabet{\mathsfit}{bold}{\encodingdefault}{\sfdefault}{bx}{n}

\newcommand{\E}{\mathbb{E}}

\DeclareMathOperator{\sign}{sign}

\DeclareMathOperator{\dist}{dist}
\newcommand{\Norm}{\mathcal{N}}
\newcommand{\vphi}{{\bm{\varphi}}}
\newcommand{\vrho}{{\bm{\rho}}}

\newcommand{\tabincell}[2]{%
  \begin{tabular}{@{}#1@{}}
    #2
  \end{tabular}
}

\title{Featurized Bidirectional GAN:\@
Adversarial Defense via Adversarially Learned Semantic Inference}

\author[1]{Ruying Bao}
\author[2]{Sihang Liang}
\author[1]{Qingcan Wang}
\affil[1]{%
  Program in Applied and Computational Mathematics, Princeton University}
\affil[2]{Department of Physics, Princeton University}
\affil[ ]{\texttt{\{rbao,sihangl,qingcanw\}@princeton.edu}}

\begin{document}

\maketitle

\begin{abstract}
Deep neural networks have been demonstrated to be vulnerable to adversarial
attacks, where small perturbations intentionally added to the original inputs
can fool the classifier. In this paper, we propose a defense method, Featurized
Bidirectional Generative Adversarial Networks (FBGAN), to extract the semantic
features of the input and filter the non-semantic perturbation. FBGAN is
pre-trained on the clean dataset in an unsupervised manner, adversarially
learning a bidirectional mapping between the high-dimensional data space and the
low-dimensional semantic space; also mutual information is applied to
disentangle the semantically meaningful features. After the bidirectional
mapping, the adversarial data can be reconstructed to denoised data, which could
be fed into any pre-trained classifier. We empirically show the quality of
reconstruction images and the effectiveness of defense.
\end{abstract}

\section{Introduction}

The existence of adversarial examples causes serious security concern about
reliability of deep neural networks (DNN). DNN may mislabel the perturbed images
with high confidence even though the perturbation is too small to be recognized
by human. Moreover, adversarial examples will often fool several models
simultaneously, even if these models have different architectures
\citep{szegedy2014intriguing}.  One possible explanation is that when
recognizing images, human usually catch high-level and semantic features, such
as the shape of the digits in MNIST dataset, which are robust under small
perturbation; DNN may easily catch low-level and weak features, such as the
gray-scale values of certain area in the images, which are non-robust when the
pixel-wise perturbation accumulates \citep{tsipras2018there}.

Most previous adversarial defense methods fall into two classes: adversarial
training and gradient masking. Adversarial training methods
\citep{szegedy2014intriguing, tramer2017ensemble, madry2017towards,
sinha2017certifiable} apply adversarial perturbations on training data online,
and feed both the clean data and the adversarial data to train the classifier,
i.e., solve a minimax game iteratively. However, it is flawed by the high
computational cost to generate adversarial examples, especially for more complex
dataset and harder attacks. Gradient masking methods modify the architecture of
the classifier such that the attacker cannot get useful gradient information of
the inputs. One example is the thermometer encoding
\citep{buckman2018thermometer} which preprocesses the input in a one hot vector,
and such discretization prevent the attacker from backpropagating through the
input to calculate the adversarial perturbation. However,
\citet{athalye2018obfuscated} shows that gradient masking methods can be
circumvented and lead to a false sense of security in defenses against
adversarial attacks.

Both of adversarial training and gradient masking methods defend adversarial
attacks by improving the classifier. We take another approach by denoising the
adversarial examples without changing the classifier \citep{meng2017magnet,
ilyas2017robust, liao2018defense}.  Our defense is motivated by human cognition
process. The fact that adversarial examples cannot fool human suggests that
human do classification based on some semantic features that are unchanged after
the perturbation. Hence, it is natural to extract those semantic features and
doing the inference solely based on semantic information.  One closely related
work is Defense-GAN \citep{samangouei2018defense}, which trains a GAN
\citep{goodfellow2014generative} to generate the manifold of unperturbed images,
then finds the nearest point on the manifold to the adversarial example as the
denoising result. While it is a novel way to leverage generative model to filter
the adversarial perturbation, it takes iterations to search the nearest point on
the manifold, which is time consuming.

\begin{figure}
  \centering
  \begin{subfigure}{.3\textwidth}
    \centering
    \includegraphics[width=\textwidth]{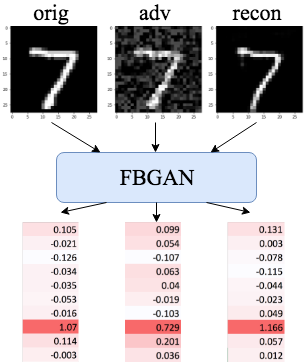}
    \caption{Semantic codes}
  \end{subfigure}
  \quad
  \begin{subfigure}{.65\textwidth}
    \centering
    \includegraphics[width=\textwidth]{fbgan.png}
    \caption{FBGAN structure}
  \end{subfigure}
  \caption{%
     (a) The semantic features of images should be unchanged before and after
     the adversarial perturbation. Via FBGAN, original, adversarial and
     reconstructed images are encoded to similar semantic codes. Each column
     stands for the ten-categorical code that related to the classification of
     an image (see \autoref{sec:fbgan} for details). Here all three images are
     classified as ``7'' from categorical codes.  (b) Besides a discriminator
     $D$ and a generator $G$ in the vanilla GAN, we add an encoder $E$ mapping
     from the data space to the latent space, and the discriminator $D$ takes a
     tuple $(\vx, \vz)$ as input. There are three types of tuple $(\vx, \vz)$:
     $(\rvx, E(\rvx))$ for $\rvx \sim P_\rvx$, $(G(\rvz), \rvz)$ for $\rvz \sim
     P_\rvz$ and $(G(E(\rvx)), E(\rvx))$ for $\rvx \sim P_\rvx$; the
     discriminator $D$ treats the first type as real and the other two as fake.
     Mutual information between latent codes $z$ and generated $G(z)$ is
     maximized in order to disentangle the semantic features.
  }\label{fig:fbgan}
\end{figure}

In this paper, we propose Featurized Bidirectional GAN (FBGAN), an encoding and
generative model that extracts the semantic features of the input images (either
original or perturbed), and reconstructs the unperturbed images from these
features. We take advantage of the generative capability of Bidirectional GAN
\citep{donahue2016adversarial, dumoulin2016adversarially}, where an encoder is
learned to map the input to its latent codes directly, instead of doing the
manifold search iterations. Inspired by InfoGAN \citep{chen2016infogan}, we
maximize the mutual information (MI) between all the latent codes and the
generated images. The MI regularization can significantly reduce the dimension
of latent space, as well as disentangle the semantic features of inputs in
different components of the latent codes, e.g., the tilt angle and stroke
thickness of digits in MNIST\@. We call the MI-enhanced latent codes as
\emph{semantic codes} (\autoref{fig:fbgan}). FBGAN is pre-trained on the clean
dataset in an unsupervised manner. With the feature-extraction and
reconstruction procedure, we can denoise the adversarial examples and fed them
into any pre-trained classifier, which shows effective defense against both
white-box and gray-box attacks (see \autoref{sec:experiments} for details).

\paragraph{Our contribution}
\begin{enumerate}
  \item FBGAN depicts a bidirectional mapping between a high-dimensional data
    space and a low-dimensional semantic latent space. We can extract the
    semantic features of the images, which is unchanged after the adversarial
    perturbation; we can also generate new images with indicated semantic
    features, such as the category and tilt angle of the digit.
  \item We denoise the adversarial example by extracting semantic features and
    reconstructing via FBGAN\@. This defense method is shown to be effective for
    any given pre-trained classifier under both white-box and gray-box attacks.
\end{enumerate}

\newpage

\section{Preliminaries}

\subsection{Generative Adversarial Networks and its derivatives}

\paragraph{Generative Adversarial Networks}
GAN \citep{goodfellow2014generative} is a generative model to learn
high-dimensional data distribution via an adversarial process. Instead of
modeling the probability density function, GAN learns a generator $G$ which is a
mapping from low-dimensional latent space $\Omega_\vz$ to high-dimensional data
space $\Omega_\vx$. Then a standard distribution (usually Gaussian) $\rvz \sim
P_\rvz$ in the latent space can be transferred into the distribution $G(\rvz)
\sim P_G$ in the data space. $P_G$ is supposed to approximate the objective data
distribution $P_\rvx$, thus a discriminator $D$ is proposed to distinguish
between samples from $P_\rvx$ and $P_G$. The generator $G$ and discriminator $D$
are represented by DNN and updated in the following minimax game:
\begin{equation}
  \min_G \max_D V_\text{GAN}(D, G) :=
  \E_{\rvx \sim P_\rvx}[\log D(\rvx)]
  + \E_{\rvz \sim P_\rvz}[\log(1 - D(G(\rvz)))].
\end{equation}
It can be shown that the theoretical optimal discriminator $D^\star$ satisfies:
\begin{equation}
  D^\star(\vx) = \frac{P_\rvx(\vx)}{P_\rvx(\vx) + P_G(\vx)}, \quad
  V_\text{GAN}(D^\star, G)
  = 2 D_\mathrm{JS}\left( P_\rvx \| P_G \right) - 2 \log 2,
  \label{eqn:gan_opt}
\end{equation}
where $P(\cdot)$ denotes the probability density of distribution $P$, and
$D_\mathrm{JS}$ is the Jensen-Shannnon divergence between two distribution. Thus
the theoretical optimal generator $G^\star$ will recover the data distribution,
i.e. $P_{G^\star} = P_\rvx$.

\paragraph{Bidirectional GAN}
BiGAN \citep{donahue2016adversarial,dumoulin2016adversarially} considers the
inverse mapping of the generator to learn the latent codes $\vz$ as feature
representation given data $\vx$. The encoder $E$ is introduced as a mapping from
data space $\Omega_\vx$ to latent space $\Omega_\vz$, and the discriminator
takes a tuple of data point and latent codes $(\vx, \vz)$ as inputs,
distinguishing between the joint distribution of $(\rvx, E(\rvx))$ and
$(G(\rvz), \rvz)$. The minimax objective becomes
\begin{equation}
  \min_{G, E} \max_D V_\textrm{BiGAN}(D, G, E) :=
  \E_{\rvx \sim P_\rvx}[\log D(\rvx, E(\rvx))]
  + \E_{\rvz \sim P_\rvz}[\log(1 - D(G(\rvz), \rvz))].
\end{equation}
The optimal condition for $D^\star$ is replacing $P_\rvx$ and $P_G$ by $P_{\rvx,
E(\rvx)}$ and $P_{G(\rvz), \rvz}$ in (\ref{eqn:gan_opt}). The optimal encoder
and generator can guarantee $G^\star(E^\star(\vx)) = \vx$ for $\vx \in
\Omega_\vx$ and $E^\star(G^\star(\vz)) = \vz$ for $\vz \in \Omega_\vz$.

\paragraph{InfoGAN}
InfoGAN \citep{chen2016infogan} is an extension of GAN that is able to learn
disentangled semantic representation. For example, one discrete latent code may
represent the class of the image while another continuous code may control tilt
angles. InfoGAN decomposes the latent codes into two parts $\vz = (\vc, \vz')$
where the semantic codes $\vc$ target the meaningful features, and noise codes
$\vz'$ which stand for incompressible noise. Then an information-theoretic
regularization is introduced to maximize MI between semantic codes $\rvc$ and
generated $G(\rvc, \rvz')$:
\begin{equation}
  \min_G \max_D V_\textrm{InfoGAN}(D, G) :=
  \E_{\rvx \sim P_\rvx}[\log D(\rvx)]
  + \E_{\rvz \sim P_\rvz}[\log(1 - D(G(\rvz)))]
  - \lambda I(\rvc; G(\rvc, \rvz')),
\end{equation}
where the mutual information $I(\rvc; \rvx) = H(\rvc) - H(\rvc | \rvx)$ and $H$
is the entropy.

\subsection{Adversarial attacks}

In the image classification task, given a vectorized clean image $\vx \in {[0,
1]}^d$, a classifier $C$ will output a label $y = C(\vx)$.  All adversarial
attacks aim to find a small perturbation $\vrho$ to fool the classifier such
that $C(\vx + \vrho) \ne y$ \citep{szegedy2014intriguing}. It can be formulated
as
\[
  \min_\vrho \|\vrho\|, \quad
  \textrm{s.t. } \vx + \vrho \in {[0, 1]}^d,\ C(\vx + \vrho) \ne y.
\]

Various attacking algorithms have been proposed to fool DNN
\citep{akhtar2018threat, papernot2016cleverhans}, and here are two most famous
attacks.

\paragraph{Fast Gradient Sign Method}

FGSM \citep{goodfellow2014explaining} is a single-step attack. Let $L(\vx, y)$
be the loss function of the classifier $C$ given input $\vx$ and label $y$. FGSM
defines the perturbation $\vrho$ as
\[
  \vrho = \varepsilon \cdot \mathrm{sign}(\nabla_\vx L(\vx, y)),
\]
where $\varepsilon$ is a small scalar. FGSM simply chooses the sign of change at
each pixel to increase the loss $L(\vx, y)$ and fool the classifier.

\paragraph{Projected Gradient Descent}
PGD \citep{madry2017towards} is a more powerful multi-step attack with projected
gradient descent:
\[
  \vx_0^\textrm{PGD} = \vx, \quad
  \vx_{t+1}^\textrm{PGD} = \Pi_\mathcal{S}\left[\vx_t^\textrm{PGD} +
  \alpha \cdot \sign\left(\nabla_x L(\vx_t^\textrm{PGD}, y)\right)\right]
\]
where $\Pi_\mathcal{S}$ is the projection onto $\mathcal{S} = \{\vx': \|\vx' -
\vx\|_\infty \le \varepsilon\}$.

\section{Featurized Bidirectional GAN}%
\label{sec:fbgan}

\subsection{Route map}

We use BiGAN framework to adversarially learn the bidirectional feature mapping,
and MI regularization to reduce the dimension of semantic codes and disentangle
the semantic features. In adversarial defense task, first we train FBGAN on
clean dataset, which is an unsupervised learning for semantic encoder $E$ and
image generator $G$. Second, given a pre-trained classifier $C$ and adversarial
data $x$, we reconstruct $x$ as $\tilde{x} = G(E(x))$ to filter the non-semantic
noise, then feed $\tilde{x}$ to the classifier and use $C(\tilde{x})$ as the
prediction.

\subsection{Formulation}

BiGAN provides a good approach to map high-dimensional image data $\vx$ to
low-dimensional latent codes $\vz = E(\vx)$, yet it has no restriction on the
semantic meaning of the latent codes $\vz$.  To eliminate the non-semantic noise
in adversarial examples, we maximize mutual information between latent codes
$\vz$ and generated $G(\vz)$.  Unlike InfoGAN where the latent codes is
decomposed into semantic codes and incompressible noise $\vz = (\vc, \vz')$ and
only $I(\rvc; G(\rvc, \rvz'))$ is maximized, here we regard all latent codes as
semantic and maximize $I(\rvz, G(\rvz))$ directly. Although the former method
may improve the diversity of the generation, our method focuses on the main
semantic features which is more robust under adversarial attack. 

To maximize the mutual information $I(\rvz; G(\rvz))$, we use Variational
Information Maximization technique. Suppose the underlying joint distribution is
$(\rvx, \rvz) \sim P$, then
\[
  I(\rvz; \rvx) = H(\rvz) - H(\rvz | \rvx)
  = H(\rvz) + \E_P [\log P(\rvz | \rvx)]
  = H(\rvz) + \max_Q \E_P [\log Q(\rvz | \rvx)],
\]
where $Q$ is taken over all possible joint distributions of $(\rvx, \rvz)$.
Assume that each semantic codes $\vz$ contain one categorical code $\evz_c$ and
$n$ continuous codes $\evz_1, \dots, \evz_n$. Assume that $Q(\cdot | \vx)$ is a
factored distribution $Q(\vz | \vx) = Q_c(\evz_c | \vx) \prod_{i=1}^n Q_i(\evz_i
| \vx)$. For the categorical code, rewrite the discrete probability $Q_c(\cdot |
\vx)$ as a vector $\vphi_c(\vx)$, i.e. ${\vphi_c(\vx)}_k = Q_c(\ervz_c = k |
\vx)$, then $\log Q_c(\vz_c | \vx) = -H(\vz_c, \vphi_c(\vx))$ where $H$ is the
cross entropy of two vectors regarding $\vz_c$ as a one-hot vector. For the
continuous codes, assume $Q_i(\cdot | \vx)$ is a Gaussian $\Norm(\varphi_i(\vx),
\sigma^2)$ for fixed variance $\sigma$. Now, define \emph{MI gap} as the
following distance
\begin{equation}
  \label{eqn:mi_gap}
  \dist(\vz, \vphi(\vx)) := -\log Q(\vz | \vx)
  = H(\vz_c, \vphi_c(\vx)) + C \sum_{i=1}^n \|z_i - \varphi_i(\vx)\|^2
\end{equation}

where $\vphi$ is the concatenation of $(\vphi_c, \varphi_1, \dots, \varphi_n)$
and $C$ is a constant. Note that the MI gap is a useful approach to maximize MI
between two variables.  

In the defense task, we want to pay more attention to the encoding $E(\vx)$ and
reconstruction $G(E(\vx))$ on given data $\vx$, and take the pair $(G(E(\vx)),
E(\vx))$ into consideration. Therefore, FBGAN has the following objective
function (as illustrated in \autoref{fig:fbgan}) {\footnotesize
\begin{multline}
  \min_{G, E, \varphi} \max_D V_\textrm{FBGAN}(D, G, E) :=
  \E_\rvx \big[\log D(\rvx, E(\rvx)) \big] \\
  + \frac{1}{2} \Big[\E_\rvz \big[\log(1 - D(G(\rvz), \rvz))\big]
  + \E_\rvx \big[\log(1 - D(G(E(\rvx)), E(\rvx))) \big] \Big]
  + \lambda \E_\rvz \dist(\rvz, \vphi(G(\rvz))).
\end{multline}}

\newpage

\subsection{Implementation}

\begin{wrapfigure}[17]{r}{.5\textwidth}
  \centering
  \vspace{-5em}
  \includegraphics[width=.5\textwidth]{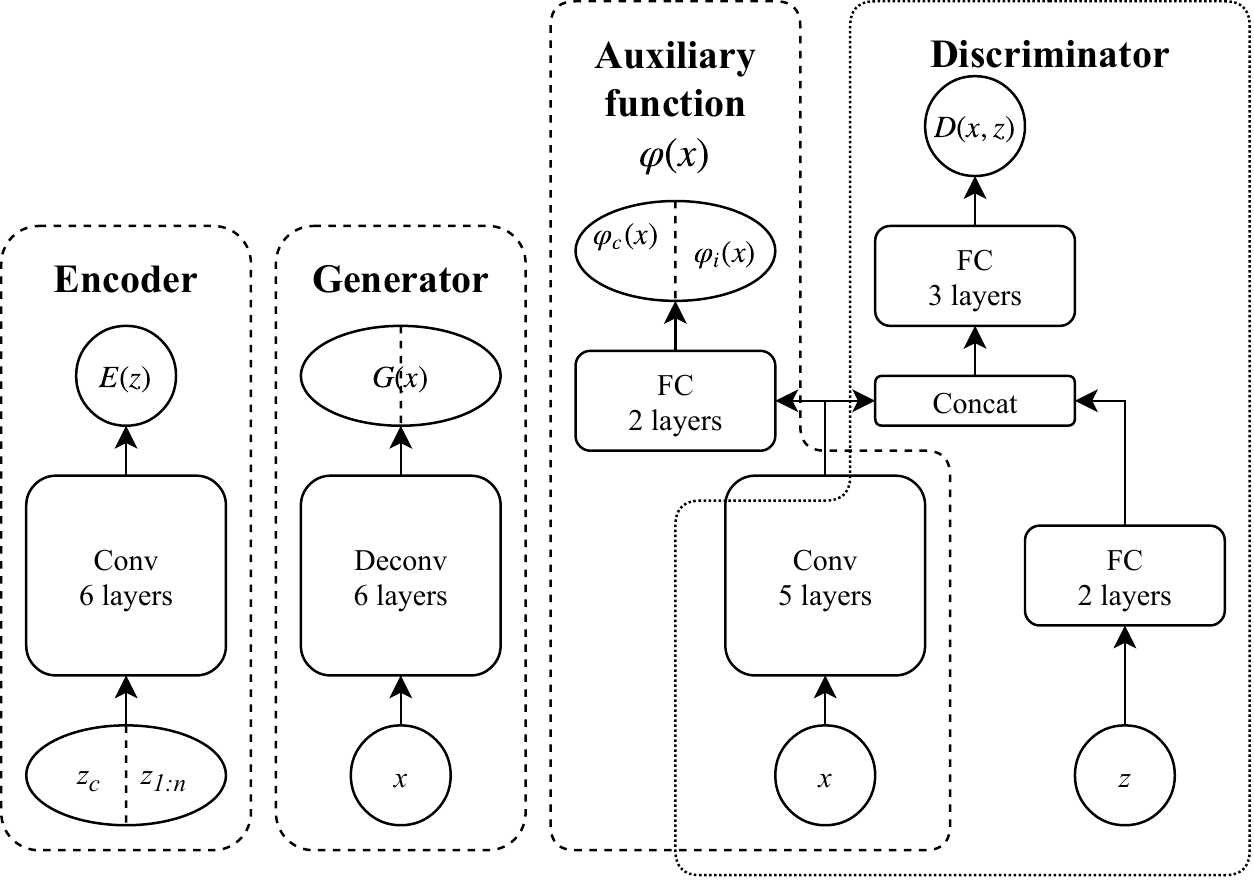}
  \caption{%
    \textbf{Implementation} \quad The encoder $E(x)$ is a convolutional network
    and the generator $G(z)$ is a deconvolutional network. The discriminator
    $D(x, z)$ shares parameters with the auxiliary function $\varphi(x)$. $\vz =
    (\evz_c, \evz_{1:n})$ stands for the categorical and continuous codes.
  }\label{fig:implement}
\end{wrapfigure}

\autoref{fig:implement} shows the implementation of FBGAN\@. $E$, $G$ and $D$
take the standard BiGAN architectures \citep{dumoulin2016adversarially}. We
replace all ReLU activation with ELU in $E$ and $G$ for smoothness, and use
weight normalization instead of batch normalization in order to ensure $E(x)$
and $G(z)$ depend only on $x$ and $z$ instead of the whole minibatch
\citep{kumar2017semi}. $E$ are trained by feature matching methods, while $G$
and $D$ are trained by the original GAN loss objectives
\citep{salimans2016improved}. The hyperparameter $\lambda = 1$. In relatively
complicated dataset such as SVHN, we add an auto-encoder term $\E_{\rvx \sim
P_\rvx} \|G(E(\rvx)) - \rvx\|^2$ in the objective function for only the last 1\%
training steps to further improve the reconstruction quality.

\section{Experiments}%
\label{sec:experiments}

We present our results in two parts: (1) Representing capability of semantic
codes. We can store the information of an image by a few number of semantic
codes, and the reconstruction from the codes keep the main features as the
original one.  (2) Defenses against gray-box and white-box attacks. In this
paper, we call \emph{gray-box} attacks as having access only to the original
classifier architectures and parameters; \emph{white-box} attacks are those have
access to both of the classifier and FBGAN details.

We focus on three datasets in our experiments: the MNIST hand-written digits
dataset \citep{lecun1998gradient}, Fashion MNIST (FMNIST) dataset
\citep{xiao2017fashion}, and the Street View House Numbers (SVHN) dataset
\citep{netzer2011reading}.

\subsection{Semantic representation}

\begin{figure}[b]
  \centering
  \begin{subfigure}{.224\textwidth}
    \centering
    \includegraphics[width=\textwidth]{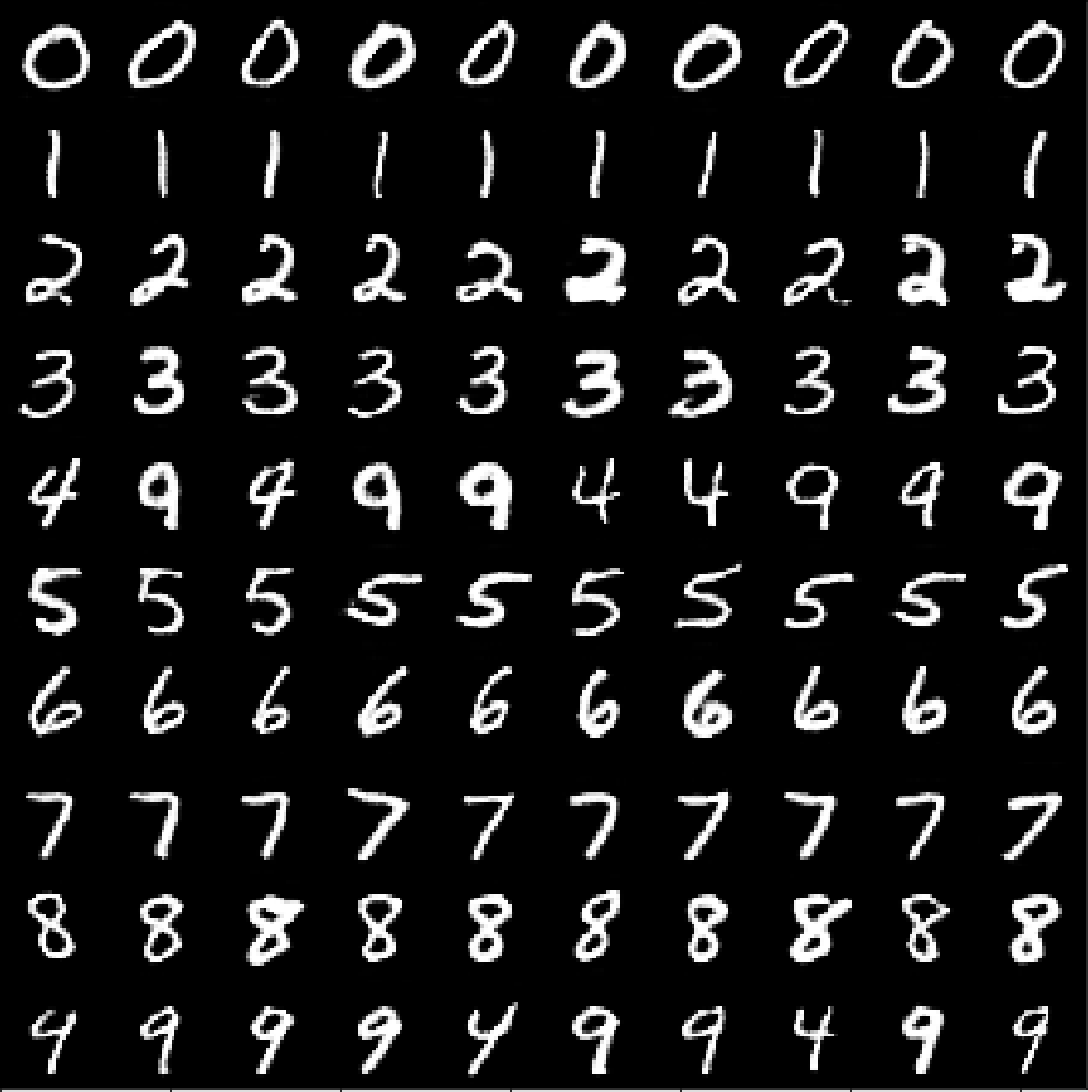}
    \caption{}
  \end{subfigure}\space%
  \begin{subfigure}{.252\textwidth}
    \centering
    \includegraphics[width=\textwidth]{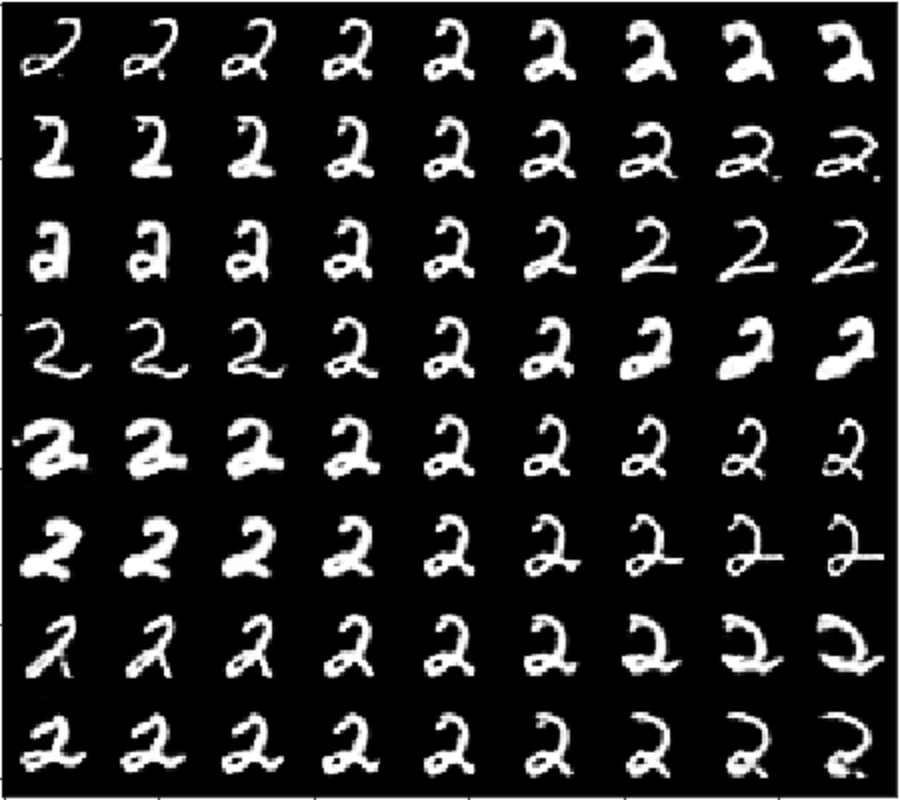}
    \caption{}
  \end{subfigure}\space%
  \begin{subfigure}{.224\textwidth}
    \centering
    \includegraphics[width=\textwidth]{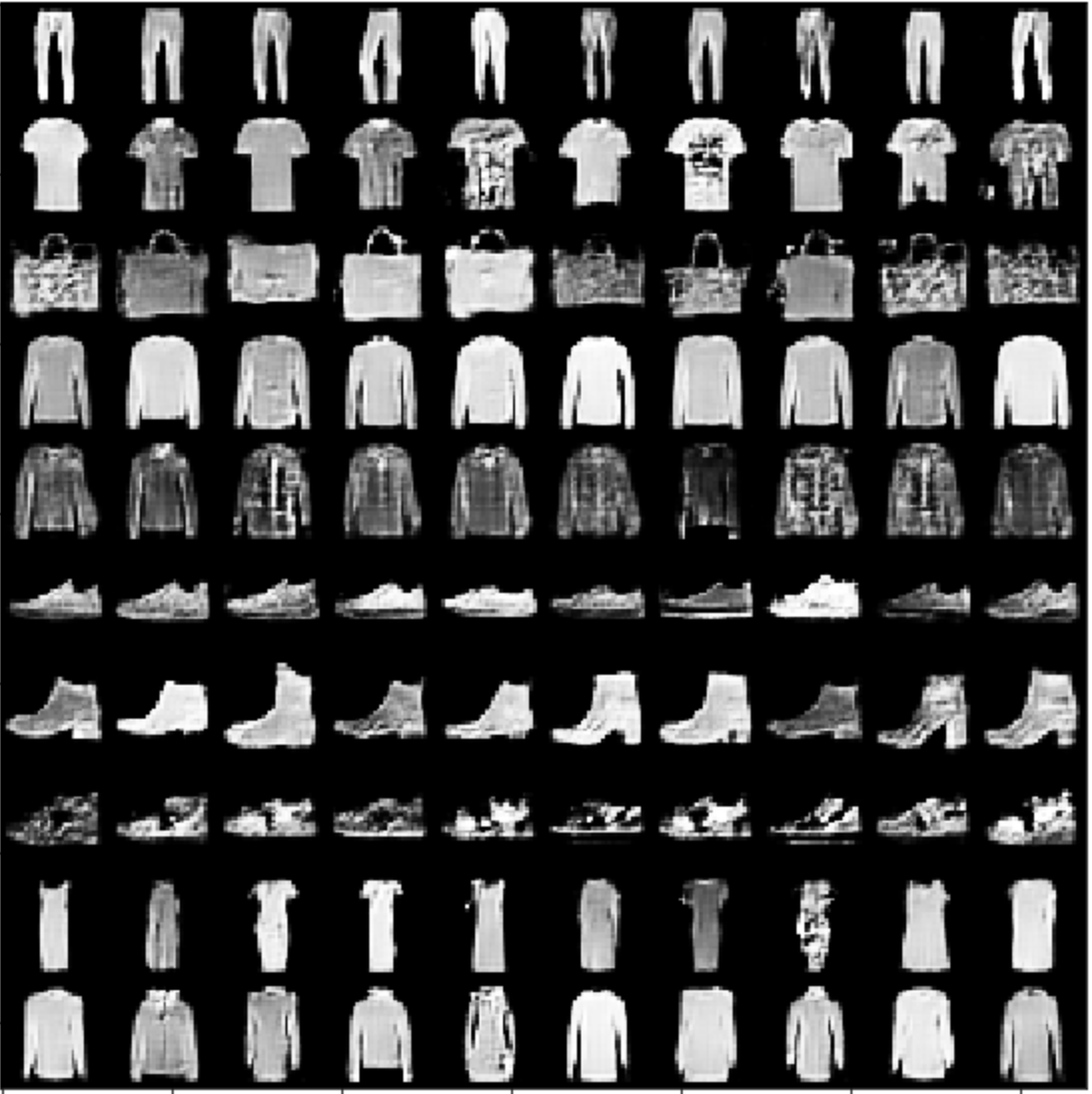}
    \caption{}
  \end{subfigure}\space%
  \begin{subfigure}{.252\textwidth}
    \centering
    \includegraphics[width=\textwidth]{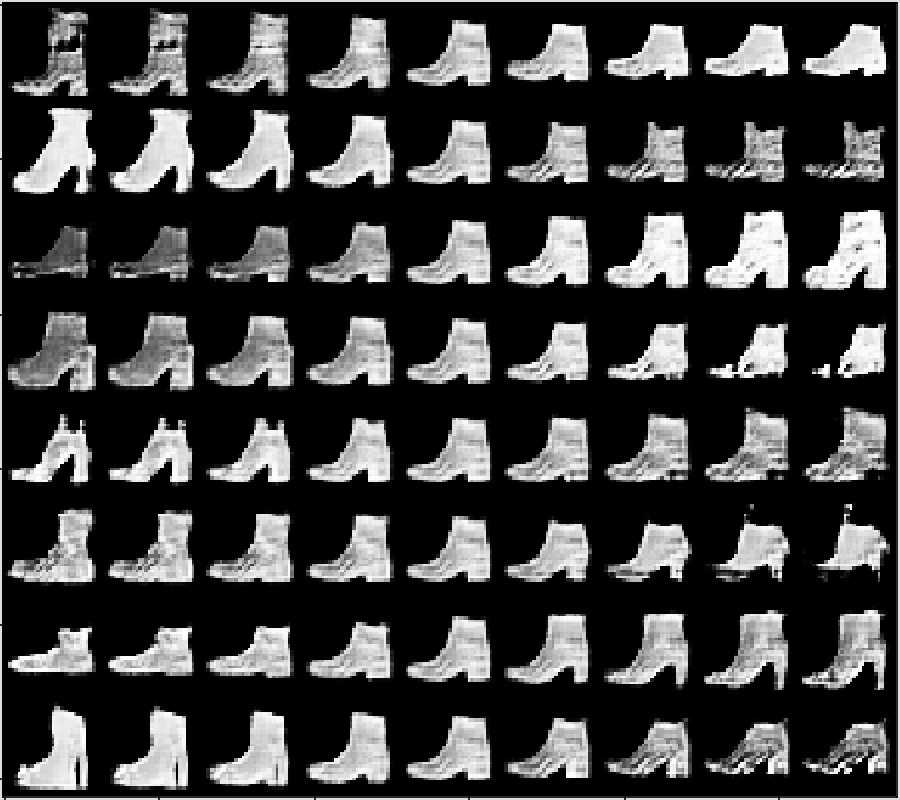}
    \caption{}
  \end{subfigure}\space%
  \caption{%
    \textbf{Manipulating semantic codes on MNIST and FMNIST} \quad Images
    generated by one ten-dimensional categorical code and eight continuous
    codes. (a) and (c) demonstrate that we can generate any category of images
    by changing the categorical codes. (b) and (d) are the effects of continuous
    codes: each row shows how the generated image changes when tuning one
    continuous codes with all other codes fixed.
  }\label{fig:mnist_generate}
\end{figure}

FBGAN can present the semantic features of MNIST by one ten-dimensional
categorical code and only four continuous codes, and FMNIST by one
ten-dimensional categorical code and eight continuous codes.  Previous related
works require much higher latent space dimension. For example in InfoGAN, one
ten-dimensional categorical code and three continuous codes and 128 random
noises codes are used.

Categorical code can learn the most significant modes in a data distribution.
For example, the ten-categorical code in MNIST / FMNIST represents ten different
digits / fashion products.  The continuous codes can finely tune the more
detailed features of a certain mode. \autoref{fig:mnist_generate} shows ten
MNIST digits generated by FBGAN and the effect of tuning different continues
codes.

We observe that the reconstruction of MNIST and FMNIST datasets are of high
qualities. The encoder first encodes a semantic representation, which is then
fed into the generator. The reconstructed image not only maintains the category,
but also detailed features as the input.

\subsection{Adversarial defenses}

\begin{table}[tp]
  \centering
  \caption{%
    Classification accuracy (\%) under different attack and defense methods for
    MNIST and FMNIST\@. The perturbation $\varepsilon$ is in $l_\infty$ norm.
    FBGAN here uses one ten-dimensional categorical code and 8 continuous codes.
    Gray-box attacks only apply to noise-filtering-type defense, and we compare
    FBGAN and Defense-GAN under the same setting. For white-box attack, the
    adversarial training with PGD $\varepsilon = 0.3$ is one of the state of the
    art results. Although better than FBGAN, adversarial training has its
    limitation: if the attack method is harder than the one used in training
    (PGD is harder than FGSM), or the perturbation is larger, then the defense
    may totally fail. FBGAN is effective and consistent for any given
    classifier, regardless of the attack method or perturbation.
  }\label{tab:mnist_reconstruct}
  \begin{tabular}{c c c | c c | c c c c}
    \toprule
    & & & \multicolumn{2}{c|}{Gray-box} & \multicolumn{4}{c}{White-box} \\
    Attack & $\varepsilon$ & \tabincell{c}{No\\ defense}
    & FBGAN & \tabincell{c}{Defense\\GAN}
    & FBGAN & \tabincell{c}{Adv train\\FGSM 0.3}
    & \tabincell{c}{Adv train\\PGD 0.1} & \tabincell{c}{Adv train\\PGD 0.3} \\
    \midrule
    \multicolumn{9}{c}{MNIST} \\
    Clean & 0   & 99.3  & 97.6  & 93.6  & 97.6  & 99.2  & 99.5  & 98.8  \\
    FGSM  & 0.1 & 78.2  & 96.6  & 95.2  & 93.4  & 97.4  & 97.9  & 97.6  \\
    FGSM  & 0.3 & 18.9  & 87.0  & 82.0  & 82.8  & 94.4  & 83.1  & 96.0  \\
    PGD   & 0.1 & 10.5  & 96.3  & 94.7  & 91.7  & 83.0  & 96.1  & 97.3  \\
    PGD   & 0.3 &  0.6  & 90.9  & 93.2  & 88.6  &  3.9  & 29.2  & 94.0  \\
    \midrule
    \multicolumn{9}{c}{FMNIST} \\
    Clean & 0   & 91.2  & 82.2  & 78.0  & 82.2  & 91.4  & 89.9  & 91.0  \\
    FGSM  & 0.1 & 24.2  & 76.3  & 52.6  & 62.7  & 82.6  & 81.0  & 75.9  \\
    FGSM  & 0.3 &  9.1  & 41.0  & 38.9  & 49.2  & 89.4  & 42.4  & 74.4  \\
    PGD   & 0.1 &  5.9  & 76.9  & 62.6  & 50.5  & 12.1  & 71.7  & 61.8  \\
    PGD   & 0.3 &  5.7  & 58.8  & 62.6  & 44.2  &  5.6  &  7.1  & 68.1  \\
    \bottomrule
  \end{tabular}
\end{table}

\begin{figure}[tbp]
  \centering
  \begin{subfigure}{.4\textwidth}
    \centering
    \includegraphics[width=.8\textwidth]{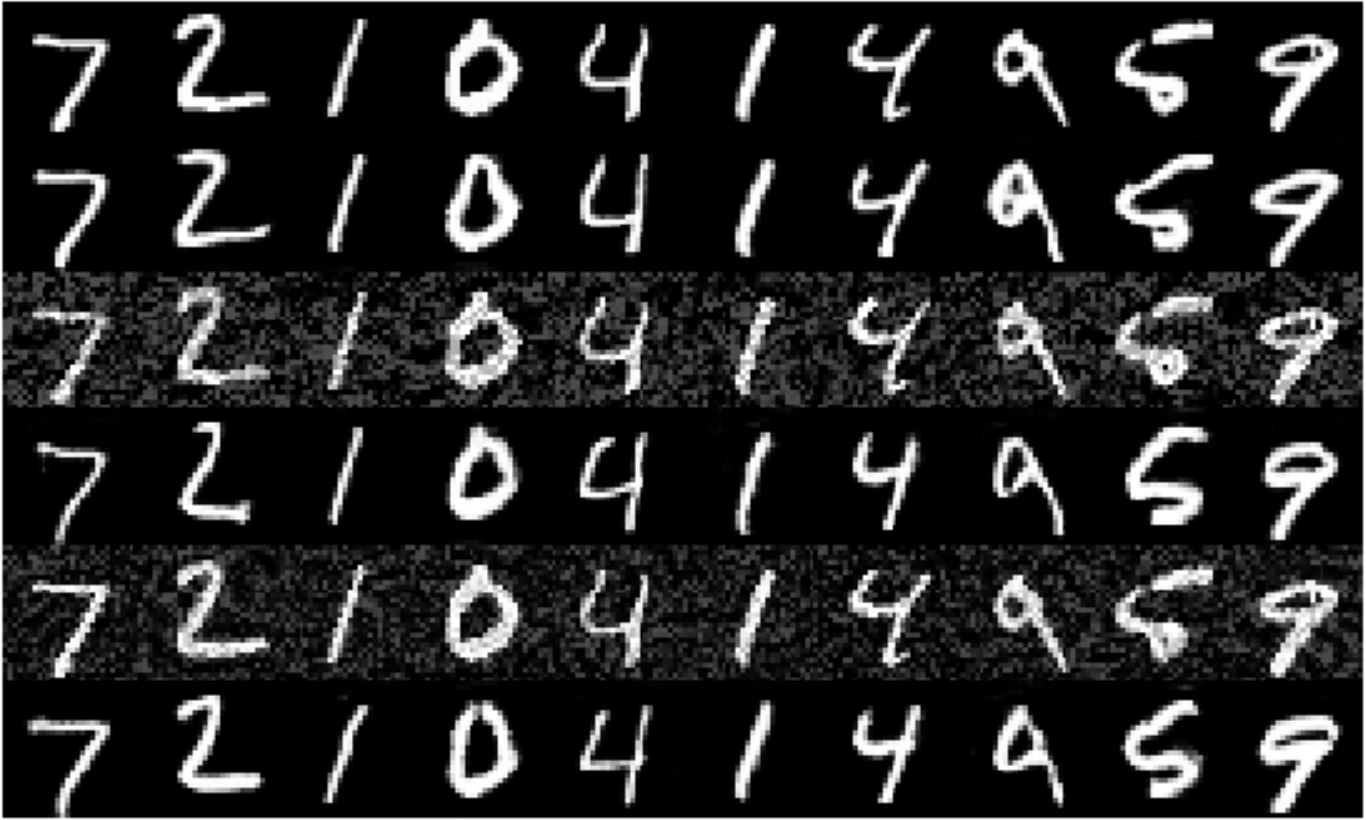}
    \caption{}
  \end{subfigure}
  \begin{subfigure}{.4\textwidth}
    \centering
    \includegraphics[width=.8\textwidth]{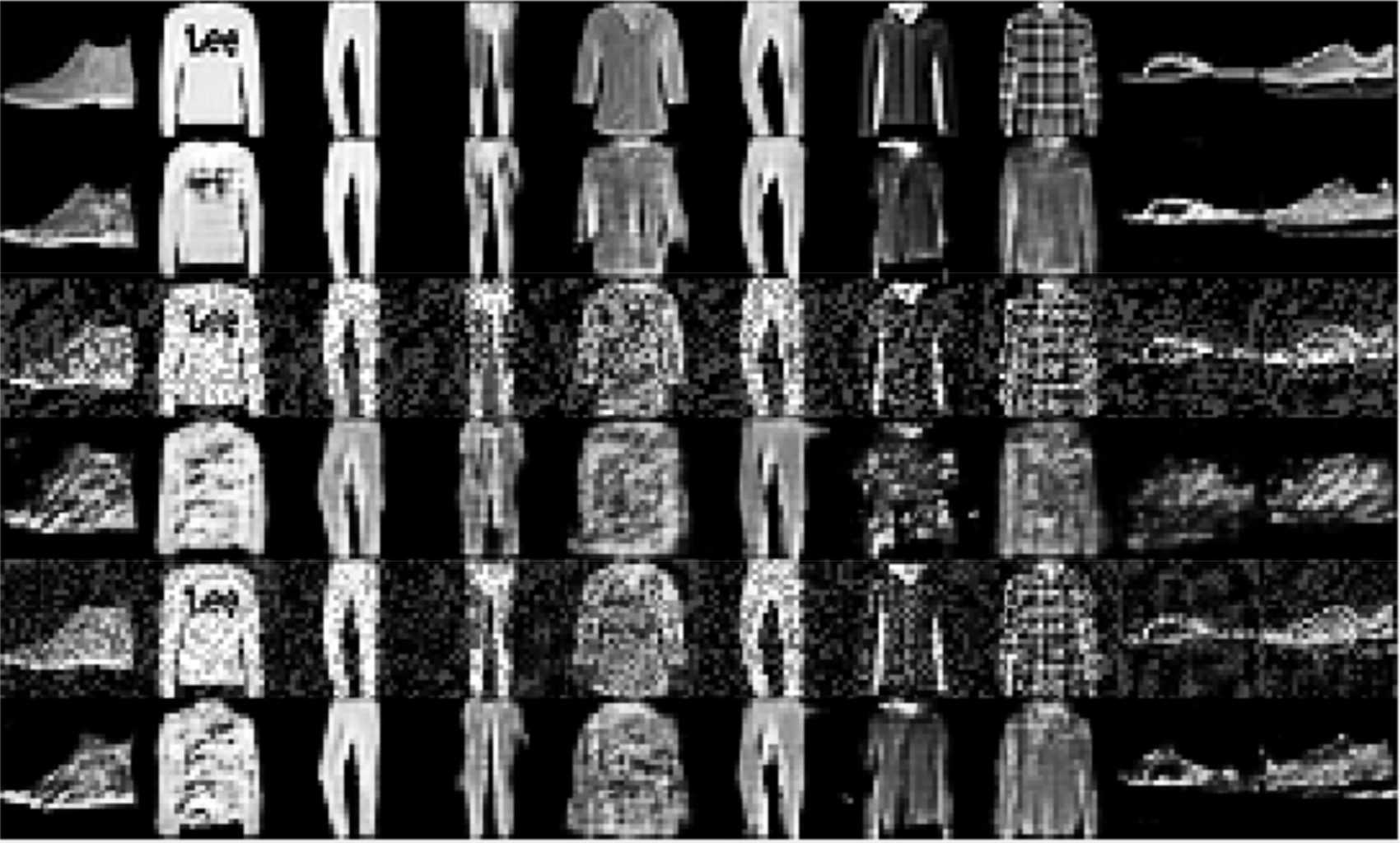}
    \caption{}
  \end{subfigure}
  \caption{%
    \textbf{Reconstruction of MNIST and FMNIST} \quad The first two rows are the
    original test set images and their reconstructions; the middle two rows are
    the gray-box adversaries and their reconstructions; the last two rows are
    the white-box adversaries and their reconstructions. All the adversaries are
    from PDG with purtabation $\varepsilon = 0.3$.
  }\label{fig:mnist_reconstruct}
\end{figure}

\begin{figure}[tbp]
  \centering
  \begin{subfigure}{.4\textwidth}
    \centering
    \includegraphics[width=.8\textwidth]{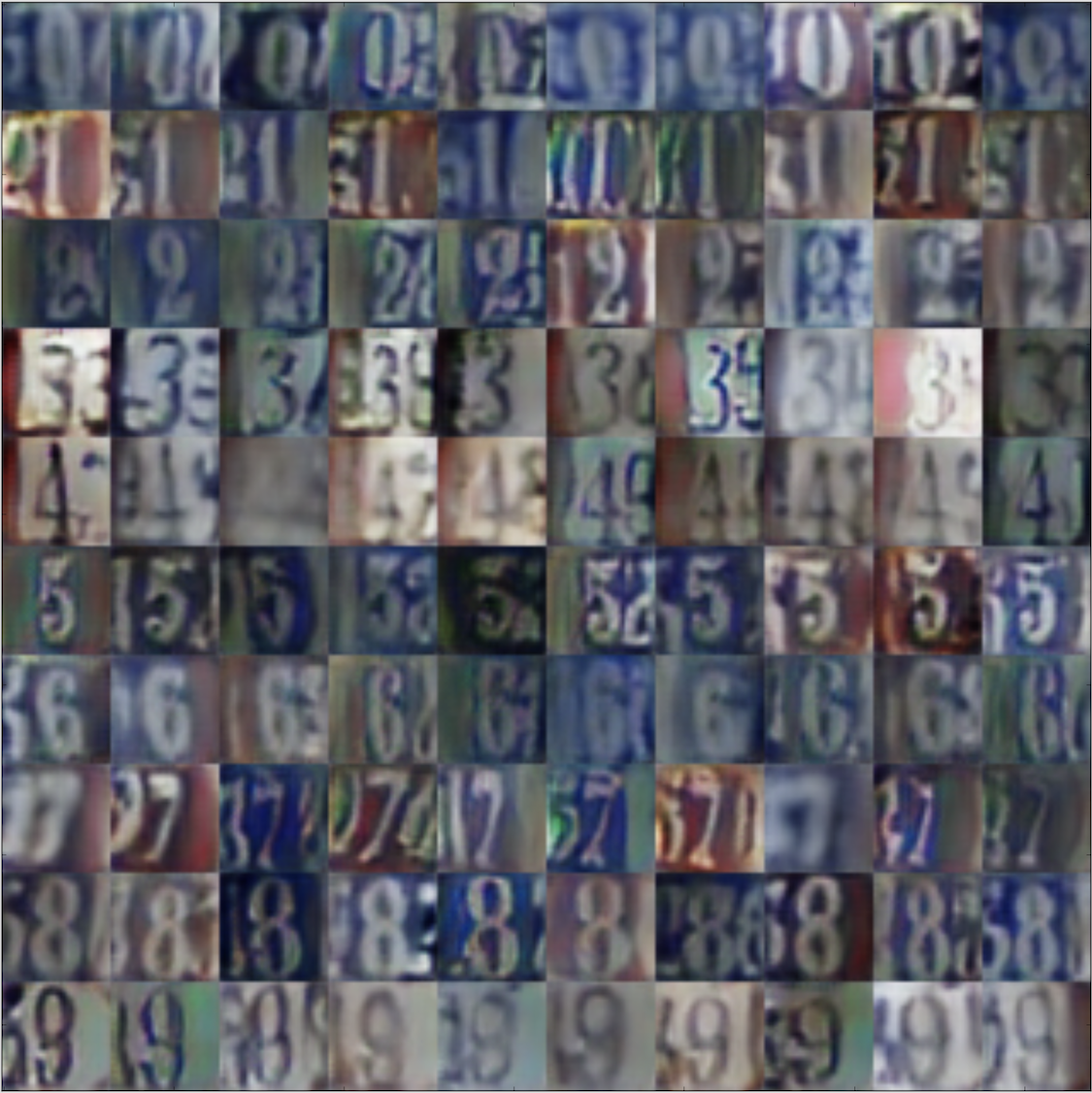}
    \caption{}
  \end{subfigure}
  \begin{subfigure}{.4\textwidth}
    \centering
    \begin{tabular}{c c c c}
      \toprule
      Attack & $\varepsilon$ & \tabincell{c}{No\\ defense} & FBGAN \\
      \midrule
      Clean & 0     & 93.7  & 83.4  \\
      FGSM  & 0.05  & 11.4  & 66.4  \\
      FGSM  & 0.10  & 10.8  & 47.7  \\
      PGD   & 0.05  &  3.4  & 71.5  \\
      PGD   & 0.10  &  2.9  & 60.9  \\
      \bottomrule
    \end{tabular}
    \caption{}
  \end{subfigure}
  \begin{subfigure}{.4\textwidth}
    \centering
    \includegraphics[width=.8\textwidth]{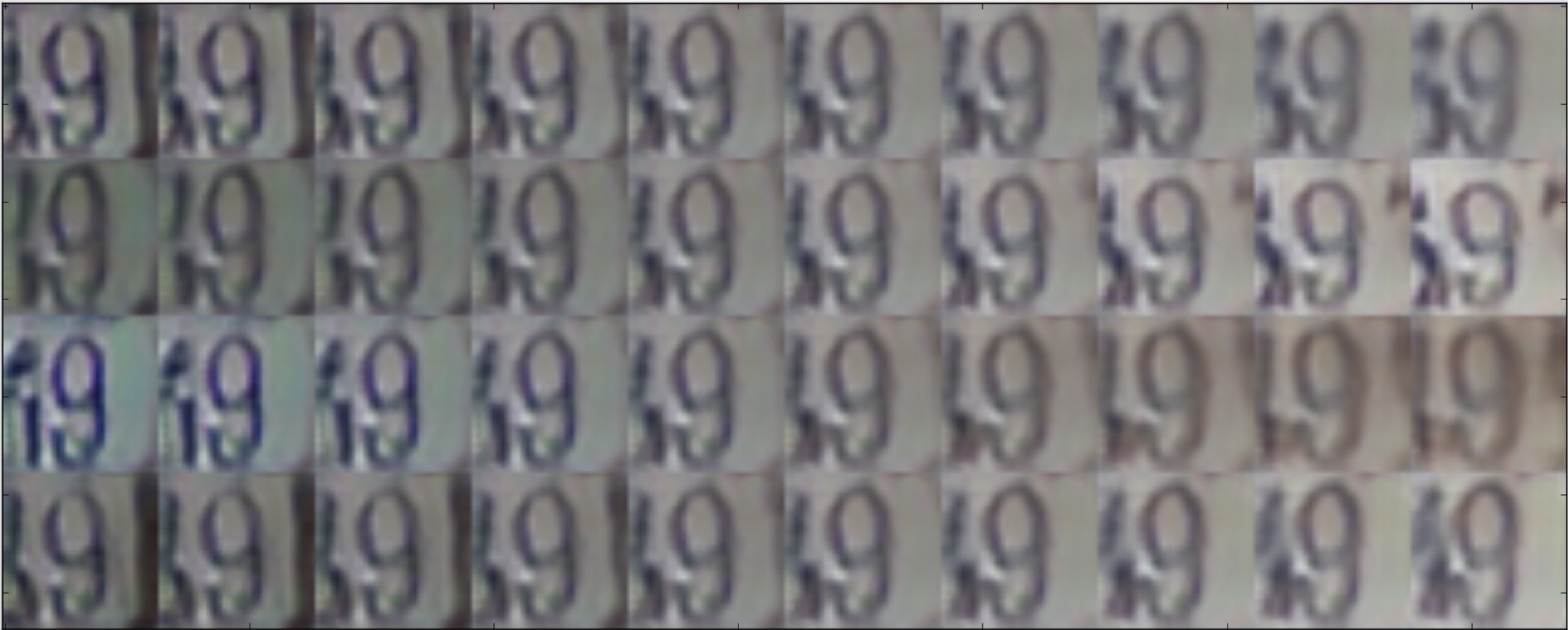}
    \caption{}
  \end{subfigure}
  \begin{subfigure}{.4\textwidth}
    \centering
    \includegraphics[width=.8\textwidth]{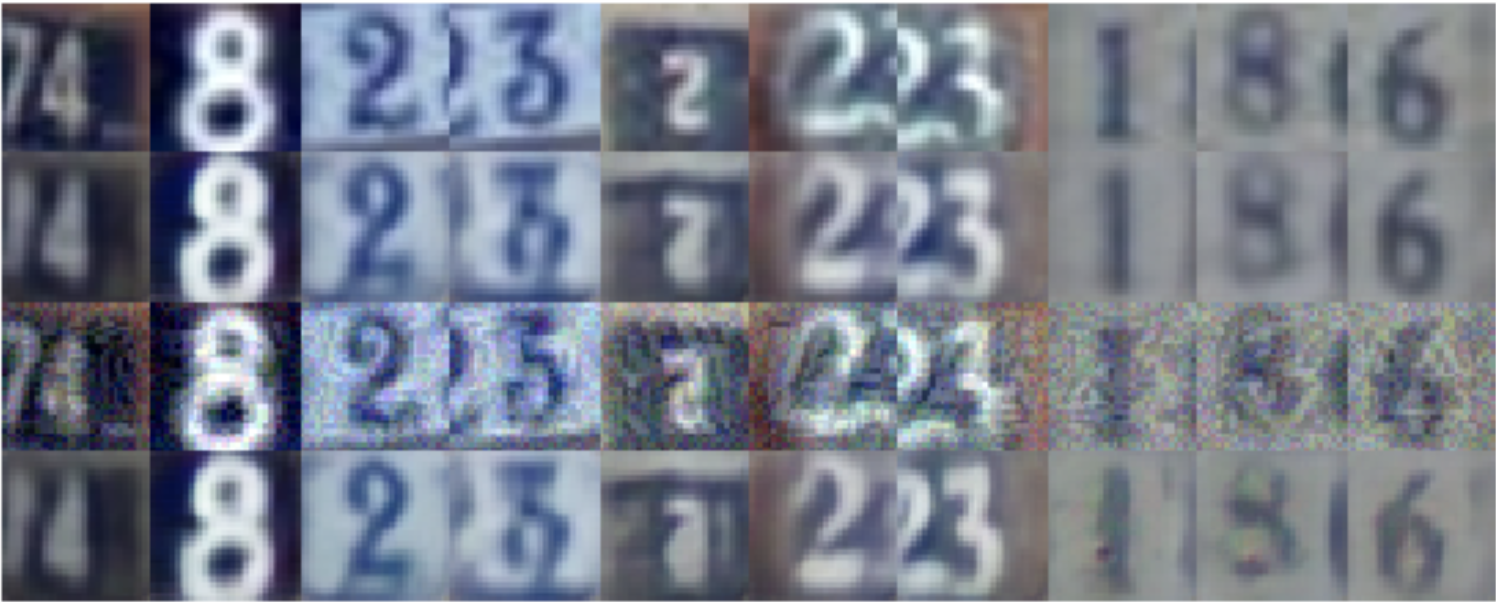}
    \caption{}
  \end{subfigure}
  \caption{%
    \textbf{Generation and reconstruction of SVHN} \quad (a) and (c) are
    generated images by changing the categorical codes and continuous codes
    respectively, similar to \autoref{fig:mnist_generate}. We observe that the
    continuous codes shown in (c) control: the blurriness (from clear to
    blurry), brightness (from bright to dark), background color (from green to
    brown) and the feature on the edge. (b) and (d) are the adversarial defense
    results. (b) shows the accuracy on clean, adversarial and reconstructed
    images, similar to \autoref{tab:mnist_reconstruct}. In (d), the first two
    rows are the clean images and their reconstructions, and the last two rows
    are the gray-box adversaries (PGD, $\varepsilon = 0.1$) and their
    reconstructions. The semantic codes consist 4 ten-categorical codes and 128
    continuous codes.
  }\label{fig:svhn}
\end{figure}

\subsubsection{Defenses against gray-box attacks}

In gray-box attacks, the attacker can only access to the classifier, but have no
information about the FBGAN filter. Hence we prepare our adversarial data by
using FGSM and PGD methods to directly attack trained classifiers. The
classifier tested on the original MNIST dataset has accuracy of 99.26\%, and the
classifier tested on the original FMNIST dataset has accuracy of 91.16\%.
\autoref{tab:mnist_reconstruct} shows our defense effect against different
methods with different $\varepsilon$ values. As shown in
\autoref{fig:mnist_reconstruct}, given adversarial examples generated by PGD
method with $\varepsilon=0.3$, we have the reconstructed images with categories
and main features maintained, and there are no more attack noises there.

\subsubsection{Defenses against white-box attacks}

In white-box case, the attacker can access not only the classifier but also the
FBGAN filter. The original data $x$ is fed through the encoder $E$, the
generator $G$ and the classifier $C$ to output $C(G(E(x)))$ as the
classification. Since $E$, $G$ and $C$ are all represented as DNN, the whole
structure is a large DNN and regraded as the objective of white-box attacks.

We implement white-box defense on MNIST and FMNIST with FBGAN having one
ten-categorical code and eight continuous codes. A regularization is added to
the encoded semantic codes $z = E(x)$: for the categorical code which is
represented by a 10-dimensional probability vector, we replace it by the
corresponding one-hot vector; for the continuous codes, we clip them between
$[-1, 1]$.  Regularizing the categorical codes can map the original input to its
counterpart in the generated space, and clipping the continuous codes is to
eliminate the influence of those low probability outliers. The results are shown
in \autoref{fig:mnist_reconstruct} and \autoref{tab:mnist_reconstruct}, where
the accuracy is above 82\% on MNIST and 44\% on FMNIST with adversarial
perturbation $\varepsilon = 0.3$.

\subsection{Comparison with BiGAN and InfoGAN}

\begin{figure}[t]
  \centering
   \begin{subfigure}{.38\textwidth}
    \centering
    \includegraphics[width=\textwidth]{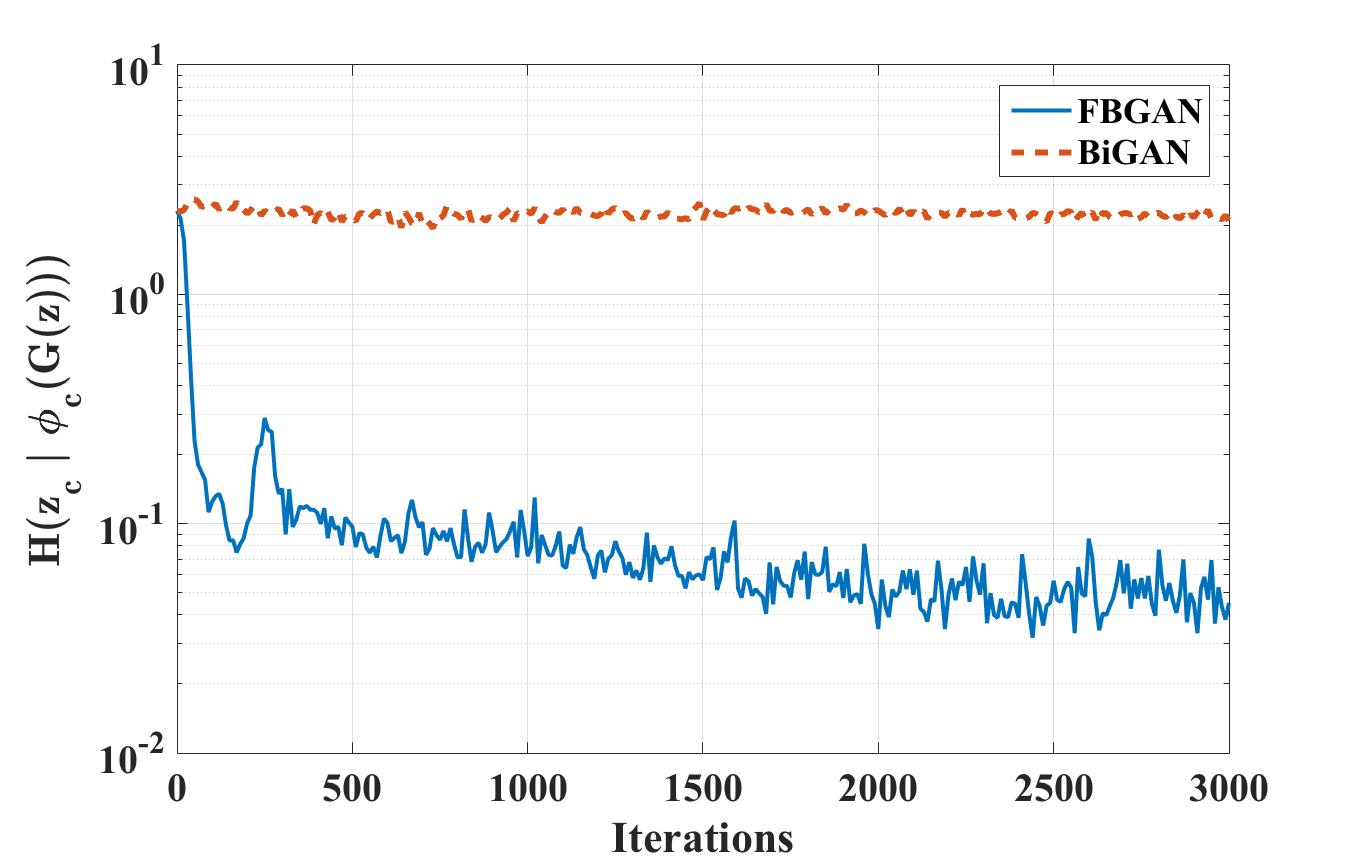}
    \caption{}
  \end{subfigure}
  \quad
  \begin{subfigure}{.24\textwidth}
    \centering
    \includegraphics[width=\textwidth]{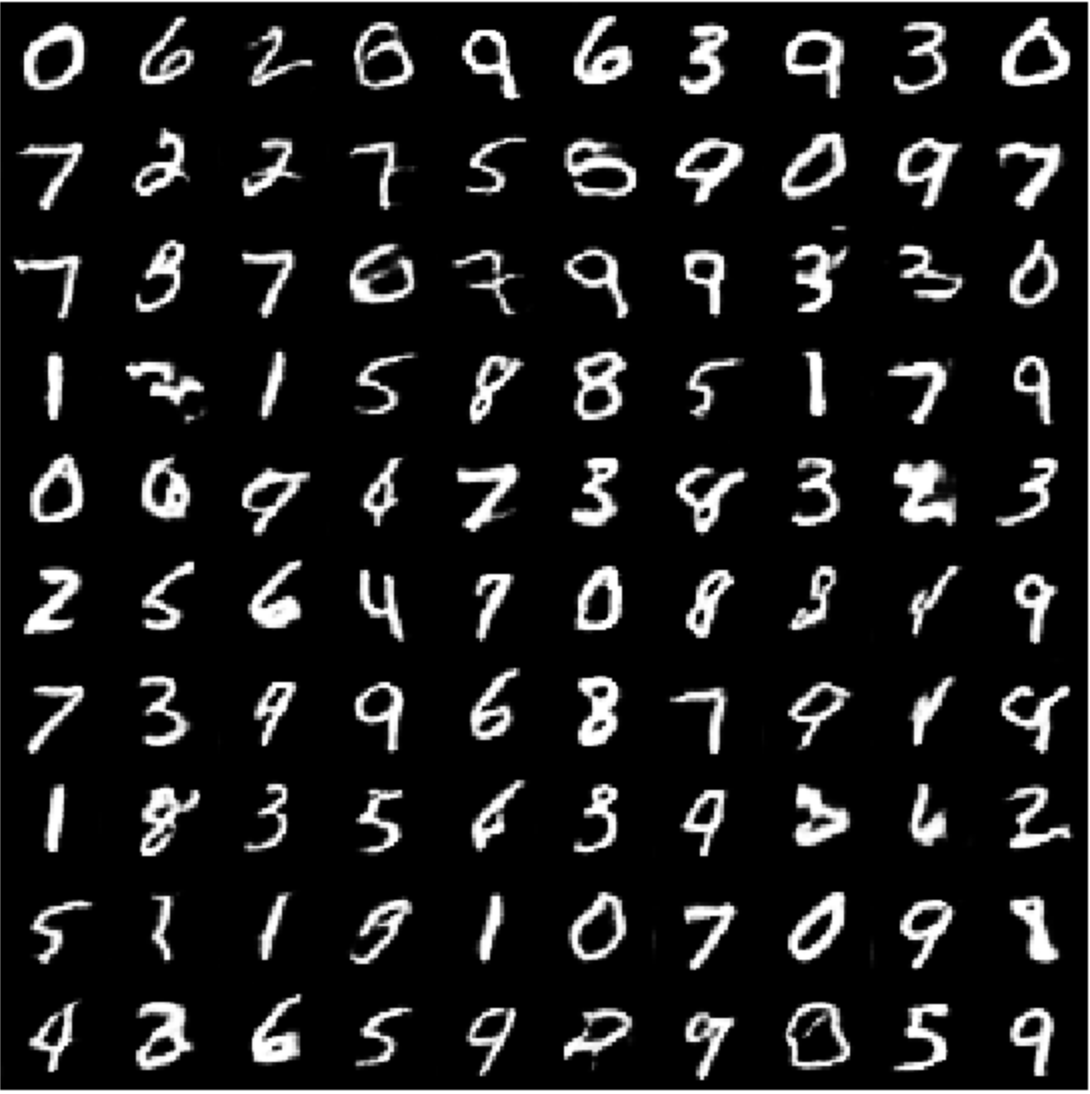}
    \caption{}
  \end{subfigure}
  \quad
  \begin{subfigure}{.27\textwidth}
    \centering
    \includegraphics[width=\textwidth]{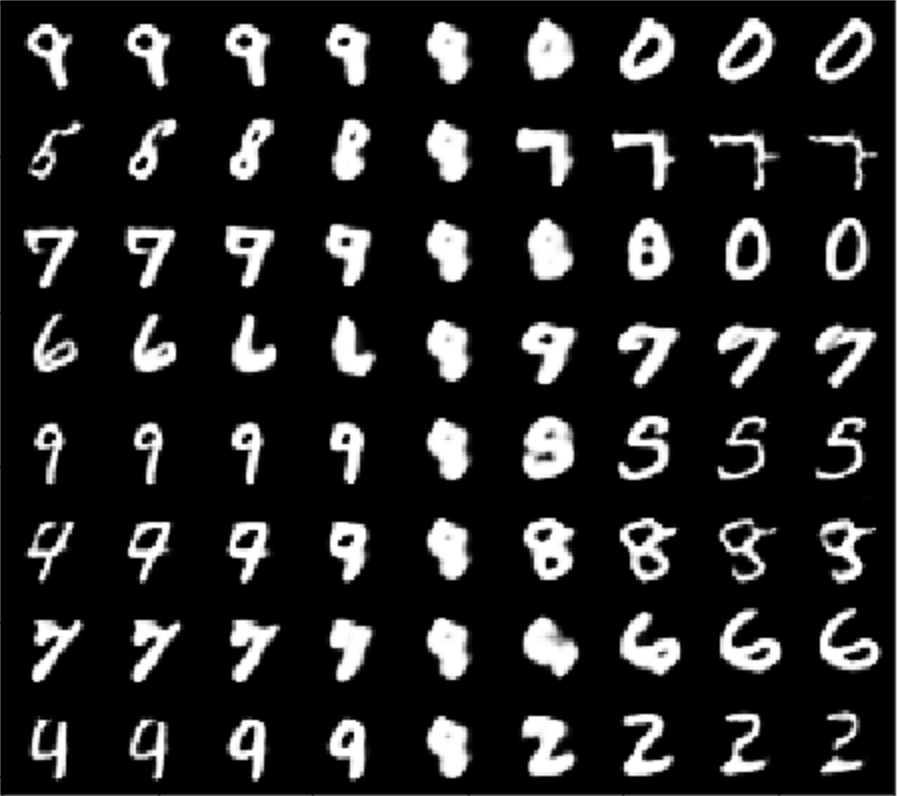}
    \caption{}
  \end{subfigure}
  \caption{%
    \textbf{Performance of vanilla BiGAN} \quad (a) illustrates MI gap
    (\ref{eqn:mi_gap}) of the categorical code, where FBGAN converges fast but
    BiGAN does not. (b) and (c) are generated images by changing the categorical
    code and continuous codes, similar to \autoref{fig:mnist_generate}. The
    semantic features are entangled in the latent codes of BiGAN\@.
  }\label{fig:bigan}
\end{figure}

BiGAN and InfoGAN are generative models aiming to produce new detailed data,
while FBGAN is a defense model aiming to regenerate data with semantic features.
The main novelty of FBGAN lies in combining the bidirectional mapping structure
and feature extraction capability for the purpose of adversarial defense. The
most important improvement from BiGAN and InfoGAN to FBGAN is the significant
reduction of the number of semantic codes by applying MI regularization on all
the semantic codes. BiGAN and InfoGAN require larger latent space to ensure the
quality and diversity of the generation, and the semantic features are stored in
latent codes in a highly entangled way; FBGAN requires much smaller latent space
to catch the basic semantic features which is robust under attacks. For example,
BiGAN and InfoGAN both employ at least 128 codes to represent and regenerate
data of MNIST, while FBGAN reduces the number to 10 categorical codes and 4
continuous codes. Hence, generative models, such as BiGAN and InfoGAN, and FBGAN
are tools for tasks in different domains.

Vanilla BiGAN without MI regularization cannot disentangle the semantic
features. Theoretically, if BiGAN achieved its optimal solution, the
minimization of JS divergence $D_\mathrm{JS}(P_{\rvx, E(\rvx)} \| P_{G(\rvz),
\rvz})$ would ensure that $H(\rvz | G(\rvz)) = 0$ and all latent codes are
effective. However, experiments show that BiGAN cannot minimize the conditional
cross entropy, and the latent codes cannot disentangle the semantic features
automatically (\autoref{fig:bigan}). Thus it is necessary to apply explicit MI
regularization.

\section{Discussion}

Nonetheless, the effectiveness of our FBGAN model against adversarial attacks
are highly dependent on the reconstruction accuracy. It is also challenging to
get a high reconstruction accuracy without over-fitting the training data. For
example, in SVHN, we apply 4 ten-dimensional categorical codes and 128
continuous codes; however, its white-box defense accuracy is much worse than
that of MNIST and FMNIST\@. We consider the various performances with different
datasets as the fact that SVHN dataset has much more modes than the rest two
datasets have. Even though the features within one category are quite different,
for example different images of number one, the background of an image adds a
large number of extra features to the object, which makes mode separation much
harder. In contrast, MNIST and FMNIST dataset with all black background could be
separated via fewer categorical codes. In our opinion, if we can find the
suitable number of categorical codes, the performance of our model will be
improved.

\bibliography{fbgan}
\bibliographystyle{plainnat}

\end{document}